\documentclass{article}
\usepackage{spconf,amsmath,graphicx,hyperref}
\usepackage{booktabs}
\usepackage{multirow}
\usepackage{subcaption}
\usepackage{amssymb}

\title{LDEPrompt: Layer-importance guided Dual Expandable Prompt Pool for Pre-trained Model-based Class-Incremental Learning}
\name{Linjie Li, Zhenyu Wu*\thanks{*Corresponding author.}, Huiyu Xiao, Yang Ji}
\address{School of Information and Communication Engineering, BUPT.\\ Engineering Research Center for Information Network, Ministry of Education}
\begin{document}
%
\maketitle
\begin{abstract}
Prompt-based class-incremental learning methods typically construct a prompt pool consisting of multiple trainable key-prompts and perform instance-level matching to select the most suitable prompt embeddings, which has shown promising results. However, existing approaches face several limitations, including fixed prompt pools, manual selection of prompt embeddings, and strong reliance on the pretrained backbone for prompt selection. To address these issues, we propose a \textbf{L}ayer-importance guided \textbf{D}ual \textbf{E}xpandable \textbf{P}rompt Pool (\textbf{LDEPrompt}), which enables adaptive layer selection as well as dynamic freezing and expansion of the prompt pool. Extensive experiments on widely used class-incremental learning benchmarks demonstrate that LDEPrompt achieves state-of-the-art performance, validating its effectiveness and scalability.  The code is available at: https://github.com/Frank-lilinjie/LDEPrompt
\end{abstract}
\section{Introduction}
\label{sec:intro}

Incremental learning (also known as continual learning or lifelong learning) aims to acquire new task knowledge without forgetting previously learned tasks, while avoiding the need to store old task data~\cite{CILsurvey}. The ultimate goal is to enable deep learning models to continuously absorb new knowledge without significantly increasing resource consumption. However, fitting new tasks inevitably disrupts the representations of previous ones, leading to the notorious problem of catastrophic forgetting~\cite{catastrophic}.

Recently, within the paradigm of pretrained models, prompt-based class-incremental learning methods have attracted increasing attention~\cite{PTMsurveyCIL}. By freezing the backbone and introducing lightweight trainable prompt vectors, these approaches achieve promising performance. Specifically, they construct a pool of trainable key-prompts, where prompts are selected for training based on their similarity to sample embeddings inferred from the frozen pretrained model~\cite{l2p,dualprompt,codaprompt,evoprompt,convprompt}. This design allows the model to maintain continual learning ability through prompt pooling and similarity-based prompt selection. Nevertheless, several critical limitations remain. First, prompts repeatedly selected across different tasks are prone to being overwritten, aggravating catastrophic forgetting. Second, the prompt pool is fixed, restricting the capacity to accommodate new tasks. Third, prompt selection depends solely on embeddings from the frozen pretrained model, which lack task-specific information and thus introduce strong dependence on the backbone. \looseness=-1

To address these issues, we propose a \textbf{L}ayer-importance guided \textbf{D}ual \textbf{E}xpandable Prompt Pool (LDEPrompt), which supports adaptive layer selection for inserting prompts and dynamic expansion of the prompt pool. Concretely, before training each new task, we perform multiple forward passes on the pretrained backbone to evaluate information gain across layers and select the most informative layers to insert prompts. To prevent overwriting prompts from previous tasks, we design a dual-pool architecture: a global prompt pool and a training prompt pool. The global pool stores prompts from all past tasks and remains frozen to preserve knowledge, while the training pool contains only task-specific prompts with a fixed capacity. During training, task-adaptive prompts are formed by selecting and extending prompts from the global pool into the training pool; after training, newly learned prompts are incorporated into the global pool. This interaction between frozen global prompts and task-specific training prompts enables dynamic pool expansion, promotes knowledge preservation, and facilitates across-task sharing. \looseness=-1

We conduct extensive experiments on widely used benchmarks, including CIFAR100~\cite{CIFAR}, CUB~\cite{CUB}, and VTAB~\cite{VTAB}. Our method consistently surpasses state-of-the-art approaches by 0.50\%, 2.91\%, and 1.37\% on the three benchmarks, respectively, demonstrating its effectiveness and scalability.

\begin{figure*}[t]
	\begin{center}
		\includegraphics[width=0.8\textwidth]{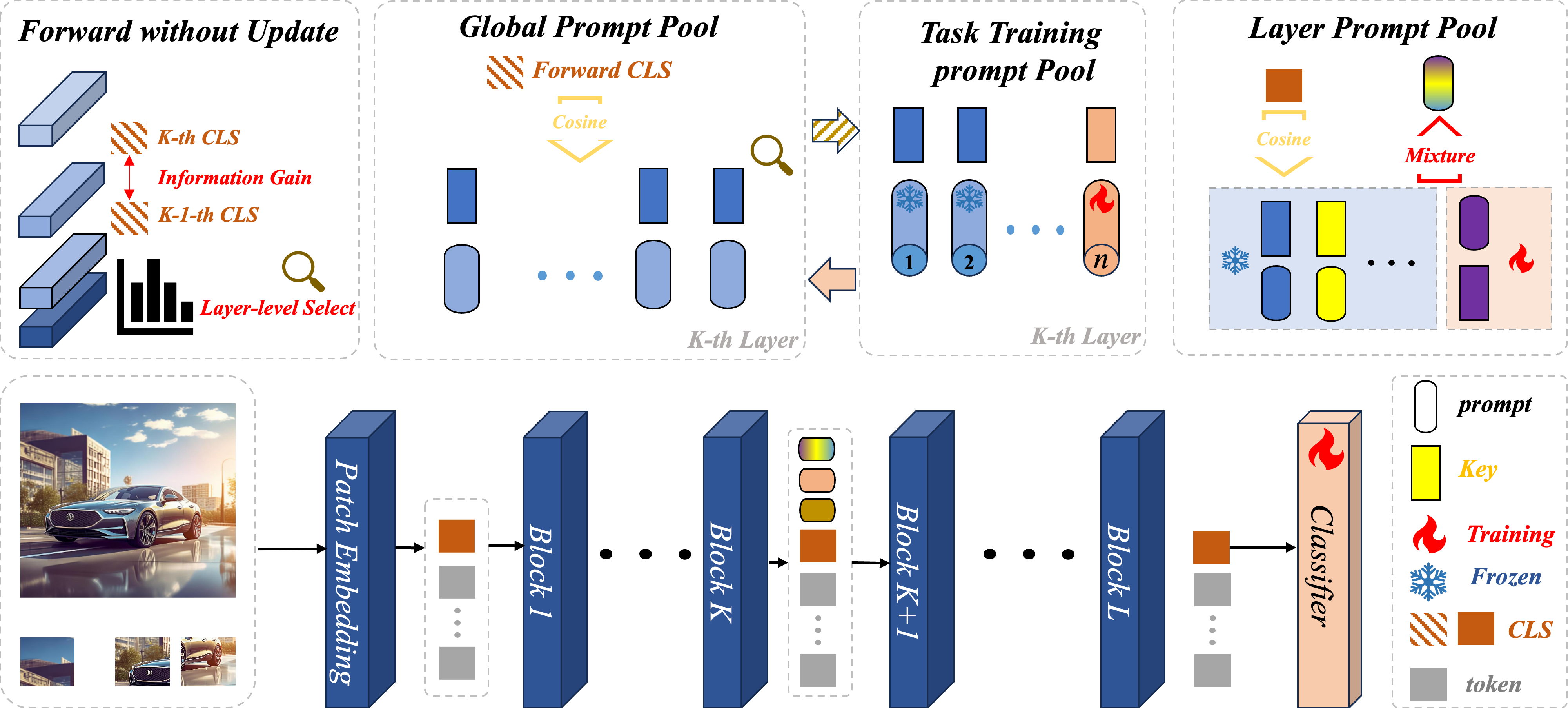}
	\end{center}
	\caption{Illustration of LDEPrompt. The upper part (from left to right) shows: (1) the forward-pass evaluation of layer importance before training; (2) task-specific prompt retrieval from the global prompt pool at layer $K$ using the CLS token; and (3) the construction of a layer-specific prompt pool of size $n$, where selected prompts are frozen while newly initialized prompts remain trainable;(4) During training, prompts within each layer are merged to form the final task-specific representations.} 
    \label{fig: overview}
\end{figure*}
\section{Preliminaries}
\subsection{Class-Incremental Learning}
Class incremental learning (CIL) aims to build a unified classifier from streaming data. In PTM-based CIL, the focus is primarily on exemplar-free CIL, where only the data from the current task is accessible when learning new tasks. For formalization, we denote the sequentially presented datasets for $t$ training tasks as $\left \{{\mathcal{D}^{1}}, {\mathcal{D}^{2}} \cdots {\mathcal{D}^{t}}\right \} $, where each $\mathcal{D}^{t}$ consists of $n_{t}$ pairs $(\mathbf{x}_{i}, \mathbf{y}_{i})$, \((\mathbf{x}, y)\) represents a data pair, where \(\mathbf{x}\) denotes the input sample and \(y\) corresponds to its associated label. In this paper, we adhere to the problem settings of previous CIL research, and cases of class overlap or multi-label co-occurrence are not within the scope of our research~\cite{CILsurvey, PTMsurveyCIL, Continualsurvey}. Therefore, the objective of CIL is to minimize the expected risk across all test datasets:
\begin{equation}
\label{Expected_Risk}
f^{\ast} = \underset{f\in \mathcal{H} }{\mathrm{argmin}} \mathbb{E} _{(\mathbf{x},\mathbf{y} )\sim \mathcal{D}^{1}\cup \cdots \mathcal{D}^{t}}\mathbb{I}(\mathbf{y}\ne f(\mathbf{x})) 
\end{equation}
Here, $f^{\ast}$ represents the optimal model, $\mathcal{H}$ is the hypothesis space, and $\mathbb{I}(\cdot)$ denotes the indicator function. $\mathcal{D}^{t}$ represents the data distribution of the task $t$, $\mathbf{y}$ represents a one-hot vector~\cite{PTMsurveyCIL,li2024tae}. 

\subsection{Prompt and Prefix tuning}
To adapt large pre-trained models, such as ViT, to downstream tasks efficiently, parameter-efficient tuning methods like Prompt Tuning~\cite{vpt} and Prefix Tuning~\cite{prefix} have been proposed. These methods avoid updating the full model by learning a small set of task-specific parameters while keeping the backbone frozen. 
Following prior work, we adopt Prefix Tuning, where learnable vectors of length $\mathit{l}_{p}$, denoted as $\mathrm{P}_{l,j}^{K} $ and $\mathrm{P}_{l,j}^{V} $, are inserted into the Key and Value matrices of the self-attention mechanism at the $j$-th layer. Specifically, the modified key and value matrices for head $h$ are given by: $\mathrm{K}_{j}^{h} =\left [ \mathrm{P}_{l,j}^{K}\mathrm{W}_{K}^{h};\mathrm{Z}_{j}\mathrm{W}_{K}^{h} \right ]$, $\mathrm{V}_{j}^{h} =\left [ \mathrm{P}_{l,j}^{V}\mathrm{W}_{V}^{h};\mathrm{Z}_{j}\mathrm{W}_{V}^{h} \right ]$, where $\mathrm{Z}_{j}\in \mathbb{R}^{(N+1)\times D} $ represents the input token sequence at layer $j$, and $\left [ \cdot ; \cdot \right ] $ denotes concatenation along the sequence dimension.
By injecting learnable prefixes only into K and V, Prefix Tuning provides a strong task-specific signal at each attention layer while leaving the original backbone untouched. 

\begin{table*}[t]
	\centering
		\begin{tabular}{@{}lcccccccc}
			\toprule
			\multicolumn{1}{l}{\multirow{2}{*}{Method}}  
			& \multicolumn{2}{c}{CIFAR B0-Inc10} 
            & \multicolumn{2}{c}{CUB B0-Inc20} 
			& \multicolumn{2}{c}{VTAB B0-Inc10} \\
			& Avg($\uparrow$) & Last($\uparrow$) 
			& Avg($\uparrow$) & Last($\uparrow$) 
			& Avg($\uparrow$) & Last($\uparrow$)
			\\
			\midrule
			L2P~\cite{l2p}	& 88.83 ± 1.39 & 83.46 ± 2.00 &77.21 ± 0.91 & 68.22 ± 0.84  &78.71 ± 6.22 & 60.57 ± 7.60   \\
                Dualprompt~\cite{dualprompt}	& 89.39 ± 1.12 & 84.58 ± 0.77 &84.18 ± 0.49 & 74.27 ± 0.49 &88.59 ± 2.28 & 77.64 ± 3.33   \\
                CODA-Prompt~\cite{codaprompt}	& 90.60 ± 0.33 & 87.01 ± 0.19 &83.89 ± 1.52 & 75.33 ± 1.06  &88.16 ± 2.10 & 76.80 ± 2.69   \\
                Aper-VPT-S~\cite{ADAM}	& 85.20 ± 0.43 & 79.77 ± 0.45 & \underline{90.21 ± 1.42} & \underline{86.23 ± 0.21}  &91.21 ± 0.46 & 83.78 ± 0.60   \\
                Aper-VPT-D~\cite{ADAM}	& 86.26 ± 0.79 & 80.28 ± 0.87 &90.06 ± 1.35 & 86.12 ± 0.11   &\underline{91.23 ± 0.62} & 83.70 ± 1.39   \\
                Evoprompt~\cite{evoprompt}	& \underline{91.20 ± 0.97} & \underline{87.63 ± 0.25} &85.25 ± 3.59 & 81.29 ± 1.00   &88.41 ± 0.77 & \underline{87.55 ± 0.39}   \\
                Cprompt~\cite{cprompt}	& 90.83 ± 1.27 & 86.38 ± 1.18 &86.84 ± 1.17 & 79.70 ± 1.33   &89.71 ± 1.35 & 80.63 ± 1.15   \\
                Convprompt~\cite{convprompt} & 90.31 ± 0.20 & 86.51 ± 1.09 &86.66 ± 0.54 & 79.64 ± 1.07   &89.71 ± 1.35 & 80.63 ± 1.15   \\
			\midrule
			LDEPrompt(ours) & \bf91.60 ± 0.80  & \bf 88.13 ± 0.76  & \bf92.31 ± 0.29 & \bf88.12 ± 0.55& \bf94.44 ± 0.11 & \bf88.92 ± 0.62  \\
			\bottomrule
		\end{tabular}
    \caption{\small Average and final accuracy comparison across three datasets. Bold numbers indicate the best performance, while \underline{underlined} values represent the second-best. All methods are evaluated under the exemplar-free setting. }
    \label{tab:benchmark}
\end{table*}

\section{Method}
\subsection{Layer Importance Evaluation}
Before training, we forward the data of the current task $D^{t}$ into the pretrained backbone model $f_{\theta }$. For each layer $l$, we compute its information gain, defined as the difference between the mutual information of the layer input and output:
\begin{equation}
IG(l)=I(x;h_l)-I(x;h_{l-1}),
\end{equation}
where $h_{l-1}$ and $h_l$ denote the representations before and after layer $l$, and $I(\cdot;\cdot)$ denotes mutual information. We then normalize the information gain across all layers using a softmax function:
\begin{equation}
\alpha _{l}=\tfrac{\mathrm{exp} (IG(l))}{ {\textstyle \sum_{j=1}^{L}} \mathrm{exp} (IG(j))} ,
\end{equation}
where $L$ is the total number of layers. Layers with importance scores above the mean are selected, and layer-specific prompt pools are attached to these positions.

\subsection{Dual Expandable Prompt Pools}
We design two prompt pools: a global prompt pool $\mathcal{P}_{g} $ and a current task training prompt pool $\mathcal{P}_{c}$. For each new task $t$, the training pool $\mathcal{P}_{c}$ is constructed with a fixed capacity $S$ and injected into the model for training. After training, the prompts are frozen and merged into the global pool $\mathcal{P}_{g} $.

When task $t>1$, different tasks may select the same layers. To enable prompt sharing across tasks without overwriting previously learned prompts, we leverage similarity matching between the intermediate features (e.g., the [CLS] token) and the stored keys in the global pool. The similarity between a feature $z$ and a prompt key $k_{i}$ is defined as:
\begin{equation}
\mathrm{sim}(z,k_{i})=\frac{z^{\top}k_{i}}{\left \| z \right \|  \left \| k_{i} \right \| }  
\end{equation}
where cosine similarity is used. For each task, we select the top-$s<S$ most similar prompts from $\mathcal{P}_{g} $, and initialize the remaining $s_{c}=S-s$ prompts randomly to form the task-specific training pool $\mathcal{P}_{c}$. After training, the newly added prompts are stored in $\mathcal{P}_{g}$, thereby expanding the global pool dynamically. \looseness=-1

\subsection{Optimization}
The overall framework of the proposed Layer-importance Guided Dual-expandable Prompt Pool (LDEPrompt) is illustrated in Figure 2. The prompts in the training pool are optimized together with the classifier using the standard cross-entropy loss:
\begin{equation}
\label{eq:CEloss}
\mathcal{L}_{ce, t} (\mathbf{x},y) = -\frac{1}{\left | \mathcal{D}^{t}  \right |} \sum_{(\mathbf{x} ,y)\in \mathcal{D}^{t}}  \mathbf{y}   \cdot \log p_{1:t}(\mathbf{x}) 
\end{equation}
where $\mathbf{y}$ is a one-hot encoded vector indicating the ground-truth class, with a value of 1 at the corresponding label position and 0 elsewhere. The term $p_{1:t}(\mathbf{x})$ represents the predicted probability distribution over all previously seen classes for the input $\mathbf{x}$.

\begin{figure*}[t]
	\centering
	\begin{subfigure}{0.3\linewidth}
		\includegraphics[width=1\columnwidth]{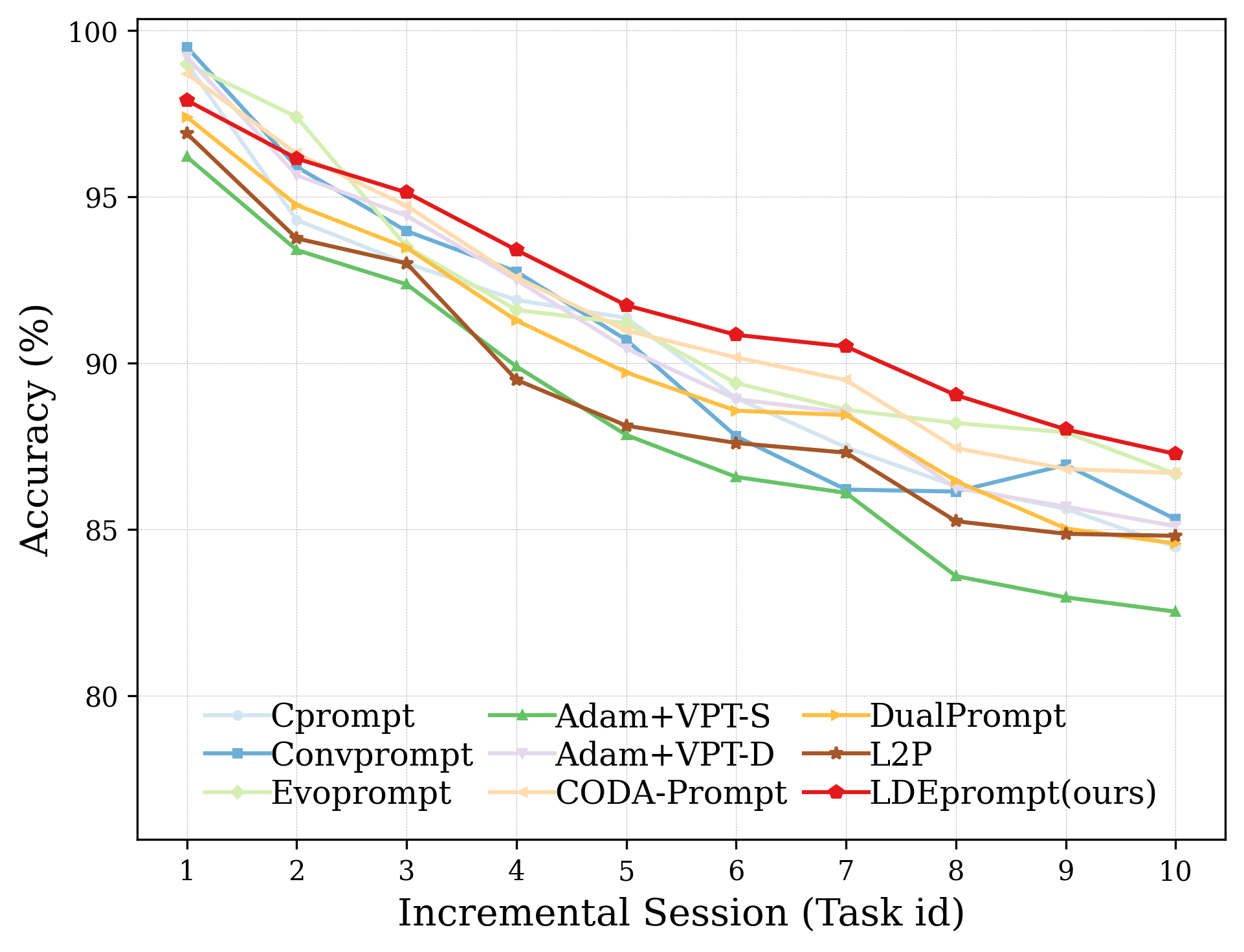}
		\caption{\small CIFAR B0-Inc10}
		\label{fig:benchmark-cifar}
	\end{subfigure}
	\hfill
	\begin{subfigure}{0.3\linewidth}
		\includegraphics[width=1\linewidth]{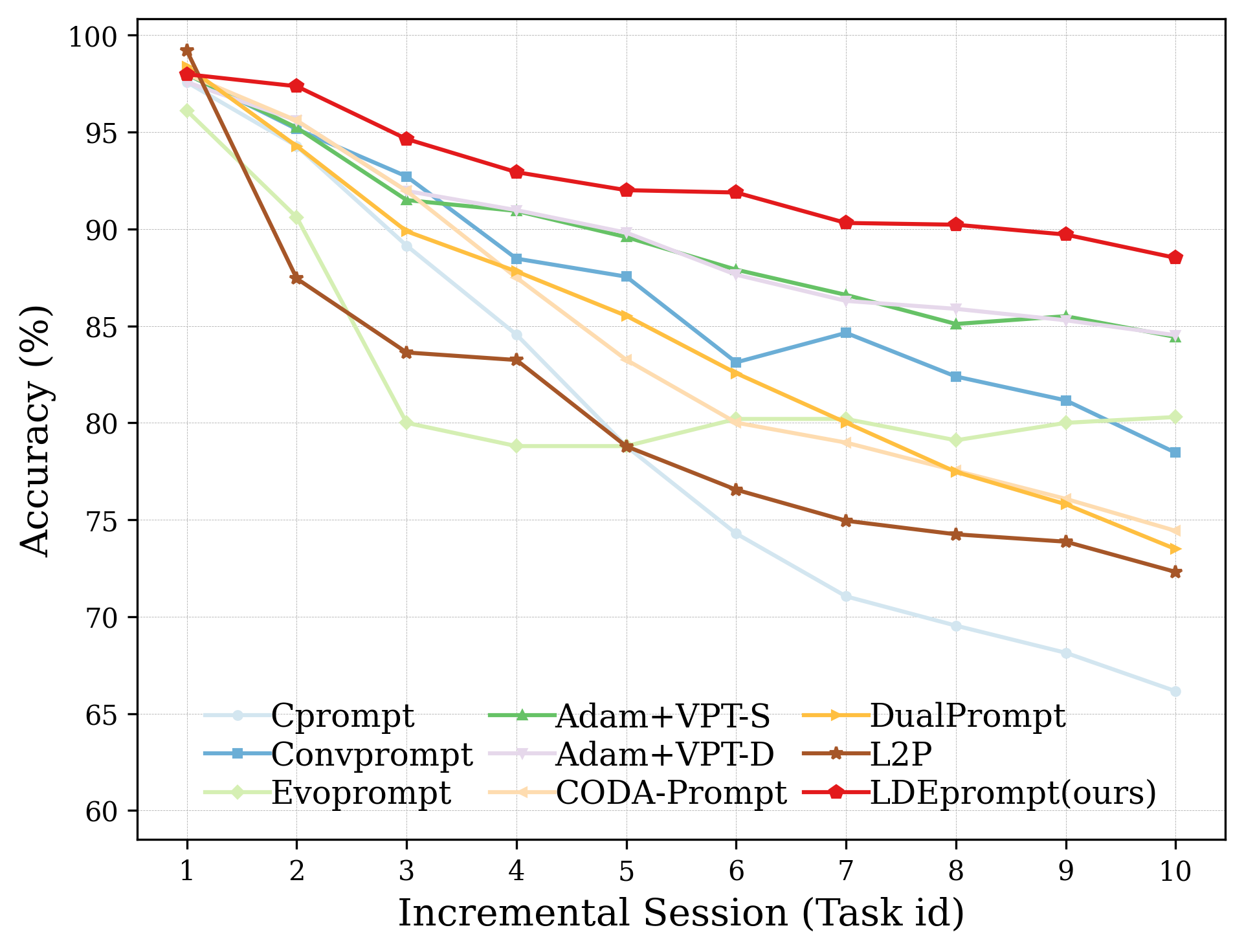}
		\caption{\small CUB B0-Inc20}
		\label{fig:benchmark-cub}
	\end{subfigure}
	\hfill
	\begin{subfigure}{0.3\linewidth}
		\includegraphics[width=1\linewidth]{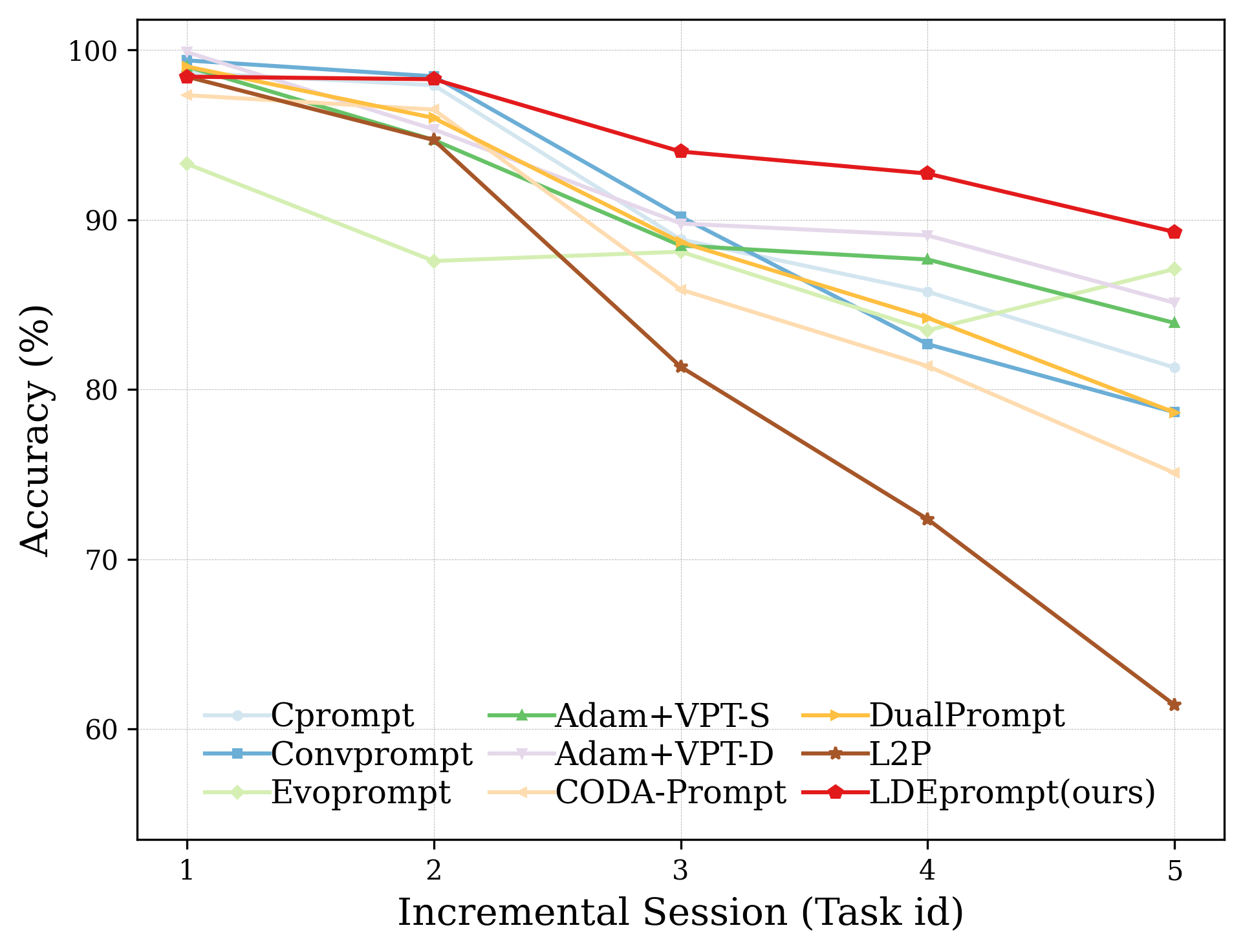}
		\caption{\small VTAB B0-Inc10}
		\label{fig:benchmark-vtav}
	\end{subfigure}
	\caption{Performance curve of various methods under different settings.}
	\label{fig:benchmarkcompare}
\end{figure*}

\section{Experiments}
\subsection{Implementation Details}
\noindent\textbf{Dataset:} We followed the research of~\cite{l2p,ADAM,li2025mote} and selected five commonly used datasets to evaluate the algorithm’s performance: CIFAR100~\cite{CIFAR}, CUB~\cite{CUB}, and VTAB~\cite{VTAB}. Specifically, CIFAR100 consists of 100 classes, CUB contains 200 classes, and VTAB includes 50 classes.

\noindent\textbf{Dataset split:} Following the benchmark setup of ~\cite{icarl,CILsurvey,ADAM}, we denote the class split as "B-$m$ Inc-$n$". Here, $m$ represents the number of classes in the first stage, and $n$ indicates the number of classes in each subsequent incremental stage, and the scope of the incremental task is $n$. We followed prior research to ensure a fair comparison and set the random seed to \{1993\}. To provide a more generalized algorithm performance evaluation, we conducted experiments on three different random seeds \{1991, 1993, 1995\} across various datasets in the main performance comparison experiments~\cite{CILsurvey}.

\noindent\textbf{Training details:} We implement our method using PyTorch~\cite{pytorch} and conduct experiments on two NVIDIA GeForce RTX 3090 GPUs. The experiments leveraged the public implementations of existing CIL methods within the PILOT framework~\cite{Pilot}. As for PTM, we follow the work of~\cite{l2p,dualprompt,ADAM,li2025mote}, we use representative models, namely {\bf ViT-B/16-IN21K} as the PTM, which is pre-trained on ImageNet21K. During training, we employ Stochastic Gradient Descent (SGD) as the optimizer with an initial learning rate of 0.001 and a weight decay of 0.005 to mitigate overfitting. The model is trained for 20 epochs with a batch size of 24 to ensure stable gradient updates.

\noindent\textbf{Evaluation protocol:} In CIL methods, we use standard metrics to evaluate performance: "Last" refers to the accuracy on the most recent task, and "Avg" refers to the Average Accuracy, which calculates the accuracy across all observed classes~\cite{CILsurvey}. Let $a_{i,j}$ be the accuracy of the model on the test set of task $j$ after training from task 1 to task $i$.

\subsection{Benchmark Comparison}
In this section, we compare LDEPrompt against several state-of-the-art (SOTA) continual learning algorithms across four widely used benchmarks. As shown in Tab.~\ref{tab:benchmark}, LDEPrompt achieves the best overall performance on three out of the four datasets: CIFAR100 B0-Inc10, CUB B0-Inc20, and VTAB B0-Inc10. Specifically, LDEPrompt surpasses the second-best method by 0.40\%, 2.11\%, and 3.23\% in average accuracy, respectively. In terms of final phase accuracy, LDEPrompt also leads by 0.50\%, 2.91\%, and 1.37\% on the same datasets.\looseness=-1

To provide a more intuitive comparison, we follow the experimental protocol using random seed 1993, and plot the accuracy at each incremental phase in Fig.~\ref{fig:benchmarkcompare}. Under this setting, LDEPrompt consistently outperforms other baselines across all incremental phases. Notably, in the final incremental stage, LDEPrompt achieves 0.57\%, 4.08\%, and 2.18\% higher accuracy than the second-best method on CIFAR100, CUB, and VTAB, respectively.\looseness=-1

\begin{table}[t]
\begin{tabular}{lccc}
\hline
\#  & CIFAR B0-Inc10 & CUB B0-Inc20   & VTAB B0-Inc10  \\ \hline
(1) & 88.23          & 87.23          & 85.34          \\
(2) & 89.03          & 89.42          & 90.12          \\
(3) & 91.56          & 91.91          & 92.87          \\
(4) & \textbf{92.00} & \textbf{92.55} & \textbf{94.55} \\ \hline
\end{tabular}
    \caption{\small Ablation results(Avg) for each component. }
    \label{tab:abla}
\end{table}

\subsection{Ablation Study}
In this subsection, we conduct ablation studies to validate the effectiveness of each component in our framework. Specifically, we compare the following variants: (1) removing the dual-pool design by using a single prompt pool without freezing or expansion; (2) retaining the dual-pool structure with prompt expansion but without freezing previous prompts; (3) inserting prompt pools into all layers; and (4) the complete LDEPrompt framework. The results in Tab~\ref{tab:abla} highlight the importance of each component. From the results of experiments (1) and (2), it is evident that both freezing and expansion are essential for effective learning. Furthermore, the comparison between experiments (3) and (4) shows that inserting prompt pools into every layer does not yield optimal performance and instead introduces parameter redundancy due to the larger number of trainable parameters. In contrast, performing layer importance evaluation in advance and selectively adding prompt pools enables the model to better adapt to downstream tasks by aligning prompt allocation with task-specific characteristics.\looseness=-1

\section{Conclusion}
In this work, we presented LDEPrompt, a layer-importance guided dual expandable prompt pool framework designed to alleviate catastrophic forgetting while supporting scalable prompt-based learning.  By leveraging information gain to adaptively select layers for prompt insertion, and by introducing a dual-pool design—consisting of a frozen global pool and a task-specific training pool—our method preserves prior knowledge while dynamically expanding the prompt space to accommodate new tasks.  The interaction between the global and training pools not only facilitates knowledge retention but also promotes cross-task prompt sharing.  Extensive experiments demonstrate that LDEPrompt achieves consistent state-of-the-art performance, surpassing existing approaches by significant margins. We believe this work highlights the importance of layer-wise prompt allocation and expandable architectures, providing a promising direction for future research in prompt-based class incremental learning. \looseness=-1

\section*{Acknowledgments}
This work was supported in part by the National Natural Science Foundation of China under Grant No. 62101061, and in part by the Cooperation Agreement Project of Kunpeng-Shengteng Science, Education, Innovation and Incubation Center, Beijing University of Posts and Telecommunications.

\vfill\pagebreak


\bibliographystyle{IEEEbib}
\bibliography{strings,refs}

\end{document}